\documentclass{article}
\pdfoutput=1

\usepackage[nonatbib, final]{neurips_2022}

\usepackage[utf8]{inputenc} 
\usepackage[T1]{fontenc}    
\usepackage{url}            
\usepackage{booktabs}       
\usepackage{amsfonts}       
\usepackage{nicefrac}       
\usepackage{microtype}      
\usepackage[table]{xcolor}         

\usepackage{wrapfig,lipsum,booktabs}
\usepackage{amsmath}
\usepackage{graphicx}
\usepackage{multirow}
\usepackage{bm}
\usepackage{bbm}
\usepackage{caption}

\title{PKD: General Distillation Framework for Object Detectors via Pearson Correlation Coefficient}

\author{%
  Weihan Cao\textsuperscript{1,3} \qquad Yifan Zhang\textsuperscript{1} 
 \thanks{Corresponding author} \qquad Jianfei Gao\textsuperscript{2} \qquad
 \textbf{Anda Cheng\textsuperscript{1,3}} \\
 \textbf{Ke Cheng\textsuperscript{1,3}} \qquad
 \textbf{Jian Cheng\textsuperscript{1}} \\
  NLPR \& AIRIA, Institute of Automation, Chinese Academy of Sciences\textsuperscript{1} \\ Shanghai AI Laboratory\textsuperscript{2} \\
  School of Artificial Intelligence, University of Chinese Academy of Sciences\textsuperscript{3} \\
  \texttt{caoweihan2020@ia.ac.cn} \\ 
  \texttt{\{yfzhang, jcheng\}@nlpr.ia.ac.cn}  \\ 
  \texttt{gaojianfei@pjlab.org.cn} \\
  \texttt{\{chenganda, chengke\}2017@ia.ac.cn} \\
}

\begin{document}

\maketitle

\begin{abstract}
Knowledge distillation(KD) is a widely-used technique to train compact models in object detection. However, there is still a lack of study on how to distill between heterogeneous detectors.
In this paper, we empirically find that better FPN features from a heterogeneous teacher detector can help the student although their detection heads and label assignments are different. However, directly aligning the feature maps to distill detectors suffers from two problems. First, the difference in feature magnitude between the teacher and the student could enforce overly strict constraints on the student. Second, the FPN stages and channels with large feature magnitude from the teacher model could dominate the gradient of distillation loss, which will overwhelm the effects of other features in KD and introduce much noise.
To address the above issues, we propose to imitate features with Pearson Correlation Coefficient to focus on the relational information from the teacher and relax constraints on the magnitude of the features. 
Our method consistently outperforms the existing detection KD methods and works for both homogeneous and heterogeneous student-teacher pairs. Furthermore, it converges faster. With a powerful MaskRCNN-Swin detector as the teacher, ResNet-50 based RetinaNet and FCOS achieve 41.5\% and 43.9\% $mAP$ on COCO2017, which are 4.1\% and 4.8\% higher than the baseline, respectively. 
Our implementation is available at https://github.com/open-mmlab/mmrazor.
\end{abstract}

\section{Introduction} \label{sec:intro}
Knowledge distillation(KD) is a widely-used technique to train compact models in object detection. However, there is still a lack of study on how to distill between heterogeneous detectors. Most previous works \cite{wang2019distilling,sun2020distilling,dai2021general,li2021knowledge,guo2021distilling} rely on detector-specific designs and can only be applied to homogeneous detectors. \cite{yang2021focal,zhixing2021distilling} conduct experiments on detectors with heterogeneous backbones, but detectors with heterogeneous detection heads and different label assignments are always omitted.  Object detection is developing rapidly and algorithms with better performance are proposed continuously. Nevertheless, it is not easy to change detectors frequently in terms of stability in practical applications. Furthermore, in some scenarios, only detectors with a specific architecture can be deployed due to hardware limitations (e.g., two-stage detectors are hard to deploy), while most powerful teachers belong to different categories. Thus, it is promising if knowledge distillation can be conducted between heterogeneous detector pairs. In addition, current distillation methods, such as \cite{yang2021focal,zhang2020improve}, usually introduce several complementary loss functions to further improve their performance, so several hyper-parameters are used to adjust the contribution of each loss function which heavily affects their abilities to transfer to other datasets. 

\begin{figure}
\centering
\includegraphics[width=\linewidth]{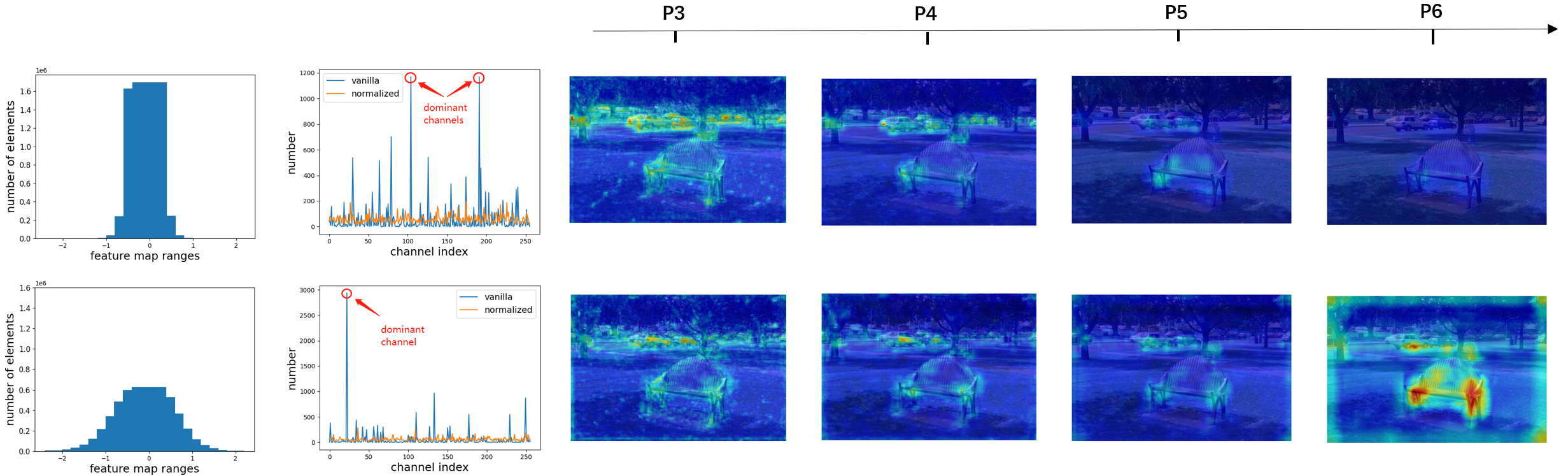}
\caption{An example to illustrate the problems suffered by directly aligning the feature maps. The first and second lines correspond to the teacher and the student. Left: Feature magnitude of the teacher and the student. Middle: Dominant channels in FPN stage 'P3'. Let $\bm{s_{l,u,v}} \in \mathbb{R}^{C}$ denote the feature vector located in pixel (u, v) from $l$-th FPN stage and omit $l$ for clarity. Then $number_i = \sum_{u, v} \mathbbm{1} [\arg \max_{c} s_{u,v}^{(c)} = i]$ where $i$ is the channel index. We define channels with a larger $number$ as dominant channels. Right: Visualization of the activation patterns from the teacher and the student. More visualization is provided in the Appendix.}
\label{fig:problem}
\end{figure}

In this paper, we first empirically verify that FPN feature imitation can distill knowledge successfully even though the student-teacher detector pairs are heterogeneous. However, directly minimizing the Mean Square Error (MSE) between features of the teacher and the student leads to sub-optimal results, with results shown in Table \ref{tab:mse}. Similar conclusions are drawn in \cite{chen2017learning,guo2021distilling,zhixing2021distilling,yang2021focal}.
In order to explore the limitations of MSE, we elaborately visualize the FPN feature responses of the teacher and student detectors, as shown in Figure \ref{fig:problem}. Specifically, for an output feature $\bm{s_l} \in \mathbb{R}^{C \times H \times W}$ from $l$-th FPN stage, we select the maximum value in the dimension $C$ at each pixel and obtain a 2-D matrix. Then we normalize the values to 0-255 according to the maximum and minimum values of $l$ 2-D matrices.
Through these comparisons, we obtain the following observations: 

\textbf{ (1) The feature value magnitudes of the teacher and the student are different, especially for heterogeneous detectors, as shown in Figure \ref{fig:problem} (left).} 
So directly aligning the feature maps between the teacher and the student may enforce overly strict constraints and do harm to the student.

\textbf{ (2) The values of several FPN stages are larger than the others, as shown in Figure \ref{fig:problem} (right).} It is obvious that FPN stage 'P6' of the teacher is less activated than FPN stage 'P3'. However, for detectors such as RetinaNet \cite{lin2017focal} and FCOS \cite{tian2019fcos}, all FPN stages share the same detection head. Hence, FPN stages with larger values could dominate the gradient of the distillation loss, which will overwhelm the effects of other features in KD and lead to sub-optimal results.

\textbf{ (3) The values of some channels are significantly larger than the others, as shown in Figure \ref{fig:problem} (middle).} However, it is mentioned in \cite{sun2020distilling,zhao2022decoupled} that the less activated features are still practical for distillation.
If they are not correctly balanced, the small gradients produced by these less activated channels can be drowned in the large gradients produced by the dominant ones, thus limiting further refinement. 
Furthermore, from the first column of Figure \ref{fig:problem} (right), we observe that there is much noise in the object-irrelevant area, because the values of the pixels in certain channels are significantly larger than those in other channels, thus being visualized in the figure. However, these pixels may be noise.
Hence, directly imitating the feature maps may introduce much noise.

\begin{figure}
\centering
\includegraphics[width=\linewidth]{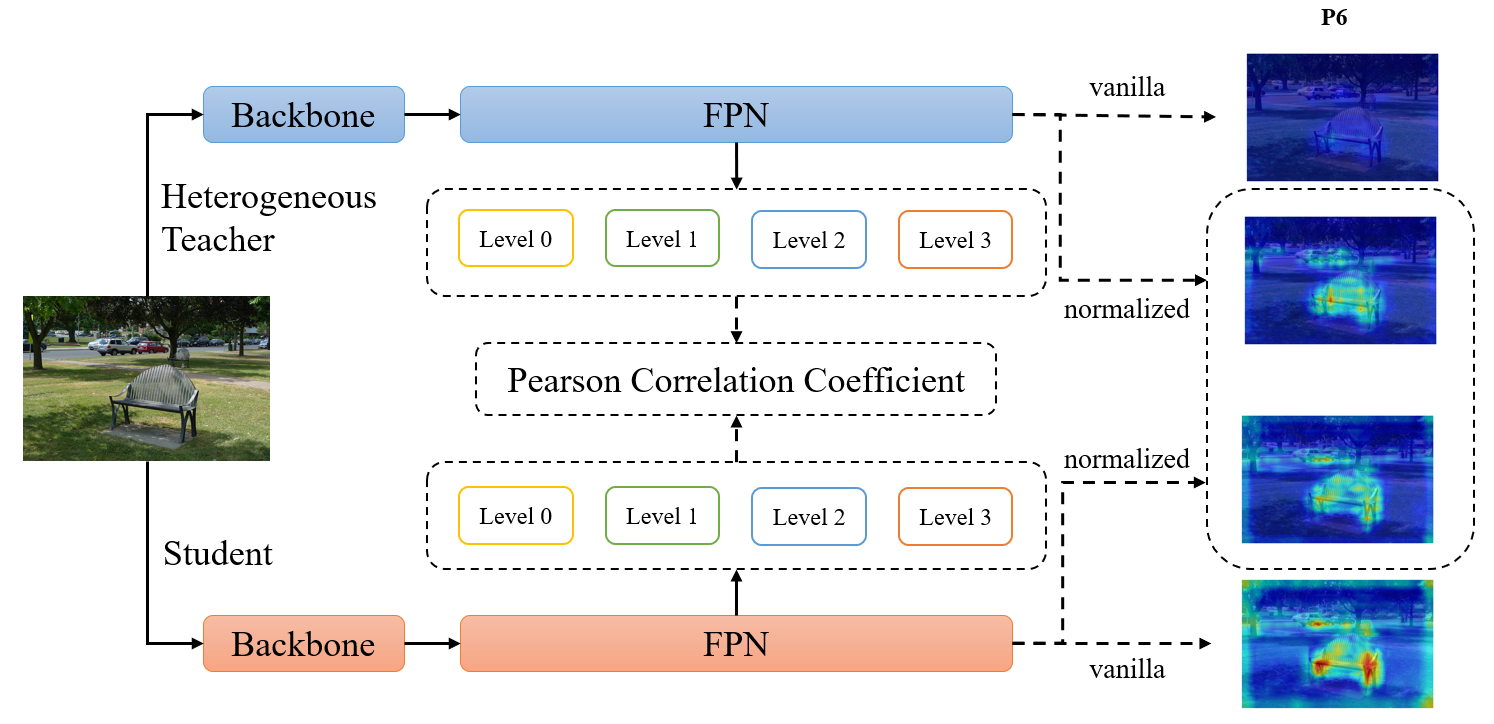}
\caption{Overview of the proposed distillation via Pearson Correlation Coefficient (PKD). To demonstrate the effectiveness of PKD, we visualize the discord activation patterns of pre-normalized and post-normalized FPN features before distillation. The normalization mechanism bridges the gap between the activation patterns of the student and the teacher.}
\label{fig:Overview}
\end{figure}

According to these observations, we propose {\bf K}nowledge {\bf D}istillation via {\bf P}earson Correlation Coefficient (PKD) shown in Figure \ref{fig:Overview}, which focuses on the linear correlation between features of the teacher and the student.
To remove the negative influences of magnitude difference between the teacher-student detector pair and within the detector among different FPN stages and channels, we first normalize the feature maps to have zero mean and unit variances and minimize the MSE loss between the normalized features. Mathematically, it is equivalent to firstly calculating the Pearson Correlation Coefficient ($r$) \cite{benesty2009pearson} between two original feature vectors, and then using $1-r$ as the feature imitation loss.

Compared with the previous methods, our method offers the following advantages. First, as the distillation loss is calculated just on FPN features, it can be easily applied to heterogeneous detector pairs, including models with heterogeneous backbones, heterogeneous detection heads and different training strategies such as label assignment. Second, as there is no need to forward the teacher's detection head, the training time can be reduced significantly, especially for those models with cascaded heads. Besides, PKD converges faster than previous methods. Last but not least, it has only one hyper-parameter - distillation loss weight and is not sensitive to it. So it can be easily applied to other datasets. We conduct extensive experiments to verify the significant performance boosts brought by our approach on COCO dataset \cite{lin2014microsoft}. Using the same two-stage detector as the teacher, ResNet50 based RetinaNet \cite{lin2017focal} and FCOS \cite{lin2017focal} achieve 41.5\% $mAP$ and 43.9\% $mAP$, which surpasses the baseline by 4.1\% and 4.8\% respectively, and it also outperforms the previous state-of-the-art methods by a large margin.

In summary, the contributions of this paper are threefold:
\begin{itemize}
\item We argue that FPN feature imitation can distill knowledge successfully even though the student-teacher detector pairs are heterogeneous. 

\item We propose to imitate FPN features with PCC to focus on relational information and relax the distribution constraints of  the student feature’s magnitude. It is capable of distilling knowledge
for both homogeneous and heterogeneous detector pairs.

\item We verify the effectiveness of our method on various detectors via extensive experiments on the COCO \cite{lin2014microsoft}, and achieve state-of-the-art performance without bells or whistles. Moreover, our method converges faster and is not sensitive to the only one hyper-parameter distillation loss weight, which is simple yet effective.

\end{itemize}

\section{Related Works}
\subsection{Object Detection}
Object detection is considered as one of the most challenging vision tasks which aims at detecting semantic objects of a certain class in images. Modern detectors are roughly divided into two-stage detectors \cite{ren2015faster,cai2019cascade,he2017mask} and one-stage detectors \cite{lin2017focal,tian2019fcos,li2020generalized,feng2021tood}. In two-stage detectors, a Region Proposal Network(RPN) is usually adopted to generate initial rough predictions refined by a task-specific detection head. A typical example is Faster R-CNN \cite{ren2015faster}. In contrast, One-stage detectors, such as RetinaNet \cite{lin2017focal} and FCOS \cite{tian2019fcos}, can directly and densely predict bounding boxes on the output feature maps. Among these works, multi-scale features are usually adopted to handle objects of various scales, {\it e.g.} FPN \cite{lin2017feature}, which is considered as a typical case for our study. The proposed PKD only distills the intermediate features and does not rely on detector-specific designs so that it can be used directly on various detectors.

\subsection{Knowledge Distillation}
Knowledge Distillation (KD) is a kind of model compression and acceleration approach aiming at transferring knowledge from a teacher model to a student model. It is popularized by \cite{hinton2015distilling} and then its effectiveness in image classification has been explored by subsequent works \cite{romero2014fitnets,huang2017like,heo2019knowledge,ahn2019variational,yim2017gift,park2019relational,liu2019knowledge,tian2019contrastive,tung2019similarity}. However, adapting KD to object detectors is nontrivial since minimizing the Kullback–Leibler (KL) divergence between the classification head outputs fails to transfer the spatial information from the teacher and only brings limited performance gain to the student. The following three strategies are usually adopted by previous methods in detection to handle the above challenge. First, the distillation is usually conducted among multi-scale intermediate features \cite{li2017mimicking}, which provides rich spatial information for detection. 
Second, different feature selection methods are proposed to overcome the foreground-background imbalance. Most of these works can be divided into three kinds according to the feature selection approach \cite{kang2021instance}: proposal-based \cite{chen2017learning,li2017mimicking,dai2021general,yao2021g}, rule-based \cite{wang2019distilling,guo2021distilling} and attention-based \cite{zhang2020improve,zhixing2021distilling,li2021knowledge}. 
Third, as the relation between different objects contains valuable information, many previous works try to improve the performance of detectors by enabling detectors to capture and make use of these relations, such as non-local modules \cite{zhang2020improve} and global distillation \cite{yang2021focal}. 

Unlike the previous works, we consider the magnitude difference, dominant FPN stages and channels as the key problem. 
We hope our method could serve as a solid baseline and help ease future research in knowledge distillation for object detectors.

\section{Method}
\subsection{Preliminaries}
In this part, we briefly recap the traditional knowledge distillation for object detection. Recently, feature-based distillation over multi-scale features is adopted to deal with rich spatial information for detection. Different imitation masks $M$ are proposed to form an attention mechanism for foreground features and filter away noises in the background. The objective can be formulated as:
\begin{equation} \label{fpn_distill}
    \mathcal{L}_{FPN}\ =\ \sum_{l=1}^{L}{\frac{1}{N_l}\ \sum_{c}^{C} \sum_{i}^{W} \sum_{j}^{H} {\ {M_{lcij}\left(F_{lcij}^t\ -\ \phi_{adapt}\left(F_{lcij}^s\right)\right)^2}}} , 
\end{equation}
where $L$ is the number of FPN layers, $l$ represents the $l$-th FPN layer, $i$, $j$ are the location of the corresponding feature map with width $W$ and height $H$. $N_l = \sum_{c}^{C} \sum_{i}^{W} \sum_{j}^{H} M_{lc ij}$. $F_{l}^{t}$ and $F_{l}^{s}$ are the $l$-th layer of feature of student and teacher detectors, respectively. Function $\phi_{adapt}$ is a 1x1 convolution layer to upsample the number of channels for the student network if the number of channels mismatches between the teacher and the student. 

The definition of $M$ is different in these methods. For example, 
FRS \cite{zhixing2021distilling} uses the aggregated classification score map from the corresponding FPN layer, and FGD \cite{yang2021focal} considers spatial attention, channel attention, object size and foreground-background information simultaneously.

\subsection{Is FPN feature imitation applicable for heterogeneous detector pairs?} \label{method:reason}

Most of the previous works perform distillation on FPN, as FPN integrates multiple layers of the backbone and provides rich spatial information of multi-scale objects. It is reasonable to force the student to imitate FPN features from its homogeneous teacher as they have the same detection head and label assignment, and better features could lead to better performance. 
However, there is still a lack of study on how to distill between heterogeneous detectors.  \cite{yang2021focal,zhixing2021distilling} conduct experiments on detectors with heterogeneous backbones, but detectors with heterogeneous detection heads and different label assignments are always omitted. Thus, we are motivated to investigate whether FPN feature imitation still makes sense for these heterogeneous detector pairs.

\begin{wraptable}{r}{5.5cm}
\caption{Results of the backbone and neck replacement on COCO.} \label{wrap-tab:1}
\centering
\vspace{-1.8em}
\begin{tabular}{lcc} \\ 
\toprule  
Backbone \& Neck & Head & $mAP$ \\ 
\midrule
FCOS-Res50 & FCOS & 36.5 \\
GFL-Res50 & FCOS & 37.6 \\ 
\midrule
Retina-Res50 & Retina & 36.3 \\
FCOS-Res50 & Retina & 35.2 \\ 
\bottomrule
\end{tabular}
\end{wraptable}

We conduct backbone and neck replacement experiments on three popular detectors: GFL \cite{li2020generalized}, FCOS \cite{tian2019fcos} and RetinaNet \cite{lin2017focal}. First, we replace the backbone and neck of FCOS with the well-trained (by 12 epochs) ones of GFL. Since the main idea of feature-based distillation methods is to directly align the feature activations of the teacher and the student, it can be considered as the extreme case of FPN feature imitation between FCOS and GFL. Then the FCOS head is finetuned with the frozen replaced GFL backbone and neck. In Table 1, it can be seen that by replacing with the backbone and neck of GFL, the detector achieves higher performance (from 36.5 to 37.6). It is verified in some extent that FPN feature imitation is applicable between heterogeneous detectors. In contrast, we replace the backbone and neck of RetinaNet with the well-trained (by 12 epochs) ones of FCOS. Due to feature value magnitude difference between the two models caused by group normalization in FCOS head, a significant mAP drop (from 36.3 to 35.2) can be observed. This means the feature value magnitude difference could interfere with the knowledge distillation between two heterogenous detectors. 


\subsection{Feature Imitation with Pearson Correlation Coefficient} \label{sec:PCC}
As discussed in Section \ref{method:reason}, a promising feature distillation approach needs to consider the magnitude difference when constructing them into pairs for imitation. Moreover, by comparing the activation patterns shown in Figure \ref{fig:problem}, we find the dominant FPN stages and channels can negatively interfere with the training phase of the student and lead to sub-optimal results, which is ignored by previous works.

To address the above issues, we propose first normalizing the features of the teacher and the student to have zero mean and unit variances and minimizing the MSE between normalized features. Additionally, we want the normalization to obey the convolutional property - so that different elements of the same feature map, at different locations, are normalized in the same way. 
Let $\mathbb{B}$ be the set of all values in a feature map across both the elements of a mini-batch and spatial locations. So for a mini-batch of size $b$ and feature maps of size $h \times w$, we use the effective mini-batch of size $m = \| \mathbb{B} \| = b \cdot hw$. Let $\bm{s^{(c)}} \in \mathbb{R}^{m}$ be the $c$-th channel of a batch of FPN outputs and omit $c$ for clarity. Then we get the normalized values $\hat{s}_{1\dots m}$ and $\hat{t}_{1\dots m}$ from the student and the teacher, respectively. Instead of delicately designing imitation masks $M$ in Eq. \ref{fpn_distill} to choose the important features, our PKD operates on the full feature map. That is, the imitation mask $M$ is filled with the scalar value 1.
Hence we can formulate our distillation loss as the following:
\begin{equation} \label{norm_mse}
\begin{aligned}
    \mathcal{L}_{FPN} = \frac{1}{2m} \sum_{i = 1}^{m} (\hat{s}_i - \hat{t}_i)^2 .
\end{aligned}
\end{equation}
Moving forward, minimizing the loss function above is equivalent to maximizing the PCC between the pre-normalized features of the student and the teacher. PCC can be computed as:
\begin{equation} \label{pcc_defination}
\begin{aligned}
    r(\bm{s}, \bm{t}) = \frac{\sum_{i = 1}^{m} (s_i - \mu_s) (t_i - \mu_t)}{\sqrt{\sum_{i = 1}^{m} (s_i - \mu_s)^2} \sqrt{\sum_{i = 1}^{m} (t_i - \mu_t)^2}} .
\end{aligned}
\end{equation}
Since $\bm{\hat{s}}, \bm{\hat{t}} \sim \mathcal{N}(0, 1)$, we get $\frac{1}{m - 1} \sum_i \hat{s}_i^2 = 1, \frac{1}{m - 1} \sum_i \hat{t}_i^2 = 1$. Then, we can reformulate Eq. \ref{norm_mse} as:
\begin{equation} \label{pcc}
\begin{aligned}
    \mathcal{L}_{FPN} &=  \frac{1}{2m} \left( (2m - 2) - 2 \sum_{i=1}^{m} \hat{s}_i \hat{t}_i \right) \\
    &= \frac{2m - 2}{2m} (1 - r(\bm{s}, \bm{t})) 
    \approx 1 - r(\bm{s}, \bm{t}) .
\end{aligned}
\end{equation}

The Pearson correlation coefficient is essentially a normalized measurement of the covariance, such that the result always has a value between $-1$ and $1$. Hence, $L_{FPN} = 1 - r$ always has a value between 0 and 2. It focuses on the linear correlation between the features of the teacher and the student, and relaxes constraints on the magnitude of the features. Actually, the feature maps $\bm{s}, \bm{t} \in \mathbb{R}^{m}$ can be regarded as $m$ data points ($s_i$, $t_i$). $L_{FPN} = 0$ implies that all data points lie on a line for which $s$ increases as $t$ increases. Hence the student is well-trained. And vice versa for $L_{FPN} = 2$. A value of 1 implies that there is no linear dependency between the features of the student and the teacher.

During training, the gradient of PCC, $\partial \mathcal{L} / \partial s_i$, with respect to each FPN output $s_i$, is given by:
\begin{equation} 
\begin{aligned}
    \frac{\partial\ \mathcal{L}_{FPN}}{\partial s_i} 
    = \frac{1}{m \sigma_s} (\hat{s_i} \cdot r(\bm{s}, \bm{t}) - \hat{t_i}) ,
\end{aligned}
\end{equation}
where $r(\bm{s}, \bm{t})$ is PCC in Eq. \ref{pcc_defination} and $\sigma_s$ is the sample standard deviation of the student's feature $\bm{s}$. 

In conclusion, PCC focuses on the relational information from the teacher and relaxes the distribution constraints on the student feature’s magnitude. Moreover, it eliminates the negative impacts of the dominant FPN stages and channels, leading to better performance. As a result, 
the normalization mechanism bridge the gap between the activation patterns of the student and the teacher (see Figure \ref{fig:Overview}). Hence, feature imitation with PCC works for heterogeneous detector pairs. We train the student detector with the total loss as follows:
\begin{equation} \label{overall}
\begin{aligned}
    \mathcal{L} = \mathcal{L}_{GT} + \alpha \mathcal{L}_{FPN} ,
\end{aligned}
\end{equation}
where $\mathcal{L}_{GT}$ is the detection training loss, $\alpha$ is the hyper-parameter to balance the detection training loss and distillation loss. 

\subsection{Connection of PCC and KL divergence} \label{sec:kl}
As discussed in Section \ref{sec:PCC}, the normalization mechanism is the key to addressing the three issues mentioned above. In previous works \cite{hinton2015distilling,wang2020intra,shu2021channel}, KL divergence has been widely used in distillation. They first convert activations into a probability distribution with softmax function and then minimize the asymmetry KL divergence of the normalized activation maps:
\begin{equation} \label{kl}
\begin{aligned}
    \mathcal{L}_{KL} = T^2 \sum_{i = 1}^{m} \phi(t_i) \cdot log[\frac{\phi(t_i)}{\phi(s_i)}] ,
\end{aligned}
\end{equation}
where $T$ is a hyper-parameter to control the degree of softness of the targets, and $\phi$ is the softmax function: 
\begin{equation} \label{softmax}
\begin{aligned}
    \phi(t) = \frac{exp(t_i / T)}{\sum_{j = 1}^{m} exp(t_j / T)} .
\end{aligned}
\end{equation}
Here, we show that minimizing KL divergence between normalized features in the high-temperature limit is equivalent to minimizing MSE between normalized ones, and hence equivalent to maximizing PCC between original ones.

Let $p_i = \phi(\hat{t_i})$ and $q_i = \phi(\hat{s_i})$ denote the probabilities from the teacher and the student, respectively. The KL divergence gradient, $\partial \mathcal{L}_{KL} / \partial \hat{s_i}$, with respect to each normalized activation $\hat{s_i}$ of the student is given by:

\begin{equation} 
\begin{aligned}
    \frac{\partial\ \mathcal{L}_{KL}}{\partial \hat{s_i}} 
    =T \left(q_i-p_i\right)
    =T \left(\frac{e^{\hat{s_i}/T}}{\sum_{j} e^{\hat{s_j}/T}}-\frac{e^{\hat{t_i}/T}}{\sum_{j} e^{\hat{t_j}/T}}\right) .
\end{aligned}
\end{equation}
If $T$ is large compared with the magnitude of the normalized activations, we can approximate:

\begin{equation} \label{kl_partial}
\begin{aligned}
    \frac{\partial \mathcal{L}_{KL}}{\partial \hat{s_i}}
    \approx T \left(\frac{1+\hat{s_i}/T}{m+\sum_{j}{\hat{s_j}/}T} - \frac{1+\hat{t_i}/T}{m+\sum_{j}{\hat{t_j}/}T}\right) .
\end{aligned}
\end{equation}
As $\hat{s_i}$ and $\hat{t_i}$ have been zero-meaned, Eq. \ref{kl_partial} simplifies to:

\begin{equation}
\begin{aligned}
    \frac{\partial \mathcal{L}_{KL}}{\partial \hat{s_i}} 
    \approx \frac{1}{m}\left(\hat{s_i} - \hat{t_i}\right) .
\end{aligned}
\end{equation}
Employing the relation between MSE and PCC shown in Eq. \ref{pcc} gives the desired result. Experimental results are provided in Section \ref{append:connection}.

\setlength{\tabcolsep}{4pt}
\begin{table}[ht]
\begin{center}
\caption{Results of the proposed method with different detection frameworks on the COCO dataset. T and S mean the teacher and student detector, respectively. 
* indicates the results reproduced by us.}
\label{table:homogeneous}
\begin{tabular}{lclccccc}
\hline\noalign{\smallskip}
Method & schedule & $mAP$ & $AP_{50}$ & $AP_{75}$ & $AP_{S}$ & $AP_{M}$ & $AP_{L}$  \\
\noalign{\smallskip}
\hline
\noalign{\smallskip}
Retina-ResX101 (T) & 2x & 40.8 & 60.5 & 43.7 & 22.9 & 44.5 & 54.6 \\
Retina-Res50   (S) & 2x & 37.4 & 56.7 & 39.6 & 20.0 & 40.7 & 49.7 \\
FKD \cite{zhang2020improve}
                   & 2x & 39.6 (+2.2) & 58.8 & 42.1 & 22.7 & 43.3 & 52.5 \\
FRS \cite{zhixing2021distilling}
                   & 2x & 40.1 (+2.7) & 59.5 & 42.5 & 21.9 & 43.7 & 54.3 \\
FGD \cite{yang2021focal}                                                                & 2x & 40.4 (+3.0) & 59.9 & 43.3 & \textbf{23.4} & 44.7 & 54.1 \\
PKD (Ours)               & 2x & \textbf{40.8 (+3.4)} & \textbf{60.3} & \textbf{43.4} & 23.0 & \textbf{45.1} & \textbf{54.7} \\
\noalign{\smallskip}
\hline
\noalign{\smallskip}
FasterRCNN-Res101 (T) & 2x & 39.8 & 60.1 & 43.3 & 22.5 & 43.6 & 52.8 \\
FasterRCNN-Res50 (S)  & 2x & 38.4 & 59.0 & 42.0 & 21.5 & 42.1 & 50.3 \\
GID \cite{dai2021general}
                  & 2x & 40.2 (+1.8) & 60.7 & 43.8 & 22.7 & 44.0 & 53.2 \\
FRS \cite{zhixing2021distilling}
                      & 2x & 40.4 (+2.0) & 60.8 & 44.0 & \textbf{23.2} & 44.4 & 53.1 \\
FGD \cite{yang2021focal}                                                                   & 2x & 40.4 (+2.0) & 60.7 & 44.3 & 22.8 & 44.5 & \textbf{53.5} \\
PKD (Ours)                  & 2x & \textbf{40.5 (+2.1)} & \textbf{60.9} & \textbf{44.4} & 22.6 & \textbf{44.8} & 53.1 \\
\noalign{\smallskip}
\hline
\noalign{\smallskip}
FCOS-Res101 (T) & 2x+ms & 41.2 & 60.4 & 44.2 & 24.7 & 45.3 & 52.7 \\
FCOS-Res50 (S)   & 2x+ms & 39.1 & 58.4 & 41.6 & 24.0 & 42.7 & 48.7 \\
FRS \cite{zhixing2021distilling} *
                & 2x+ms & 42.2 (+3.1) & 60.6 & 45.6 & \textbf{27.1} & 46.5 & 53.0 \\
FGD \cite{yang2021focal}  * & 2x+ms & 42.3 (+3.2) & 60.8 & 45.8 & 26.1 & 46.7 & 53.3 \\
PKD (Ours)            & 2x+ms & \textbf{42.8 (+3.7)} & \textbf{61.4} & \textbf{46.2} & 25.9 & \textbf{47.2} & \textbf{54.6} \\
\noalign{\smallskip}
\hline
\noalign{\smallskip}
RepPoints ResNeXt101 (T)  & 2x & 44.2 & 65.5 & 47.8 & 26.2 & 48.4 & 58.5 \\
RepPoints Res50 (S)       & 2x & 38.6 & 59.6 & 41.6 & 22.5 & 42.2 & 50.4 \\
FGD \cite{yang2021focal}  & 2x & 41.3 (+2.7) & - & - & \textbf{24.5} & 45.2 & 54.0 \\
PKD (Ours)                & 2x & \textbf{42.3 (+3.7)} & \textbf{63.1} & \textbf{45.4} & 23.9 & \textbf{46.6} & \textbf{56.5} \\
\noalign{\smallskip}
\hline
\noalign{\smallskip}
TOOD-ResX101 (T) & 2x+ms & 47.6 & 68.5 & 51.6 & 30.6 & 51.4 & 59.7 \\
TOOD-Res50 (S)   & 1x    & 42.4 & 59.7 & 46.2 & 25.4 & 45.5 & 55.7 \\
PKD (Ours)       & 1x    & \textbf{45.5 (+3.1)} & \textbf{62.8} & \textbf{49.3} & \textbf{27.4} & \textbf{49.8} & \textbf{58.4} \\
\hline
\end{tabular}
\end{center}
\end{table}
\setlength{\tabcolsep}{1.4pt}

\section{Experiments} \label{sec:experiment}
In order to verify the effectiveness and robustness of our method, we conduct experiments on different detection frameworks on the COCO \cite{lin2014microsoft} dataset. We choose the default 120k images split for training and 5k images split for the test. We use the standard training settings following \cite{yang2021focal} and report mean Average Precision ($AP$) as an evaluation metric, together with $AP$ under different thresholds and scales, $i.e.$, $AP_{50}$, $AP_{75}$, $AP_{S}$, $AP_{M}$ and $AP_{L}$.

For distillation, the hyper-parameter $\alpha$ is set to 6 when using a two-stage detector as the teacher and 10 when using a one-stage one. Feature levels of heterogeneous detector pairs may not be strictly aligned, {\it e.g.}, FasterRCNN constructs the feature pyramid from $P2$ to $P6$, while RetinaNet uses $P3$ to $P7$. To address the above issue, we upsample the low-resolution feature maps to have the same spatial size as the high-resolution ones. In addition, the adaptive layer ($\phi_{adapt}$ in Eq. \ref{fpn_distill}) is unnecessary for our PKD.

All experiments are performed on 8 Tesla A100 GPUs with two images in each. Our implementation is based on mmdetection \cite{chen2019mmdetection} and mmrazor \cite{2021mmrazor} with Pytorch framework \cite{paszke2017automatic}. 
'1x' (namely 12epochs), '2x' (namely 24 epochs) and '2x+ms' (namely 24epochs with multi-scale training) training schedules are used.
More details are given in the Appendix.

\subsection{Main Results}
Our method can be applied to different detection frameworks easily, so we first conduct experiments on five popular detectors, including a two-stage detector (Faster RCNN\cite{ren2015faster}), two anchor-based one-stage detector (RetinaNet \cite{lin2017focal}, RepPoints \cite{yang2019reppoints} and TOOD \cite{feng2021tood}) and an anchor-free detector (FCOS \cite{tian2019fcos}). Table \ref{table:homogeneous} shows the comparison of the results of state-of-the-art distillation methods on the COCO.
Our distillation method surpasses other state-of-the-art methods. All the student detectors gain significant improvements in $AP$ with the knowledge transferred from teacher detectors, {\it e.g.}, FCOS with ResNet-50 gets a 3.7\% $mAP$ improvement on the COCO dataset. These results indicate the effectiveness and generality of our method in both one-stage and two-stage detectors.

\subsection{Distilling More Student Detectors with Stronger Heterogeneous Teachers}
Most of the current methods are designed for homogeneous detector pairs. PKD is general enough to distill knowledge between both homogeneous and heterogeneous detector pairs. Here we conduct experiments on more detectors and use stronger heterogeneous teacher detectors, as shown in Table \ref{table:heterogeneous}. Comparing with Table \ref{table:homogeneous}, we find that student detectors perform better with stronger teacher detectors, {\it e.g.}, Retina-Res50 model achieves 41.5\% and 39.6\% $mAP$ with Mask RCNN-Swin \cite{liu2021Swin} and Retina-Res101, respectively. Results show that mimicking the feature maps of stronger heterogeneous teacher detectors can further boost the student's performance when applied with our PKD.

\setlength{\tabcolsep}{4pt}
\begin{table}[ht]
\begin{center}
\caption{Results of more detectors with stronger heterogeneous teacher detectors on the COCO dataset.}
\label{table:heterogeneous}
\begin{tabular}{lclccccc}
\hline\noalign{\smallskip}
Method & schedule & $mAP$ & $AP_{50}$ & $AP_{75}$ & $AP_{S}$ & $AP_{M}$ & $AP_{L}$  \\
\noalign{\smallskip}
\hline
\noalign{\smallskip}
Mask RCNN-Swin (T)    & 3x+ms & 48.2 & 69.8 & 52.8 & 32.1 & 51.8 & 62.7 \\
Retina-Res50 (S)  & 2x    & 37.4 & 56.7 & 39.6 & 20.0 & 40.7 & 49.7 \\
PKD (Ours)              & 2x    & \textbf{41.5 (+4.1)} & \textbf{60.6} & \textbf{44.1} & \textbf{22.9} & \textbf{45.2} & \textbf{56.4} \\
\noalign{\smallskip}
\hline
\noalign{\smallskip}
Mask RCNN-Swin (T)    & 3x+ms & 48.2 & 69.8 & 52.8 & 32.1 & 51.8 & 62.7 \\
FCOS-Res50 (S)   & 2x+ms & 39.1 & 58.4 & 41.6 & 24.0 & 42.7 & 48.7 \\
PKD (Ours)              & 2x+ms & \textbf{43.9 (+4.8)} & \textbf{62.3} & \textbf{47.5} & \textbf{27.2} & \textbf{48.0} & \textbf{57.1} \\
\noalign{\smallskip}
\hline
\noalign{\smallskip}
GFL-Res101 (T)    & 2x+ms & 44.9 & 63.1 & 49.0 & 28.0 & 49.1 & 57.2 \\
FCOS-Res50 (S)   & 2x+ms & 39.1 & 58.4 & 41.6 & 24.0 & 42.7 & 48.7 \\
PKD (Ours)              & 2x+ms & \textbf{43.5 (+4.4)} & \textbf{61.9} & \textbf{47.1} & \textbf{26.5} & \textbf{47.7} & \textbf{55.7} \\
\hline
\end{tabular}
\end{center}
\end{table}
\setlength{\tabcolsep}{1.4pt}

\setlength{\tabcolsep}{4pt}
\begin{table}[ht]
\begin{center}
\caption{Results of other feature-based distillation methods with the normalized feature maps.}
\label{featurebased}
\begin{tabular}{cccccc}
\hline
\noalign{\smallskip}
\multirow{2}*{FitNet} & \multirow{2}*{GT Mask} & \multirow{2}*{FRS feature imitation} & \multirow{2}*{FGD Global} & \multirow{2}*{PKD} & \multirow{2}*{\shortstack{GFL101 (44.7) \\ GFL50 (40.2)}}\\
&&&&& \\
\hline
\noalign{\smallskip}
$\checkmark$ &&&&& 41.3 \\
\rowcolor{gray!10}
$\checkmark$ &&&& $\checkmark$ & 43.3 \\
& $\checkmark$ &&&& 41.4 \\
\rowcolor{gray!10}
& $\checkmark$ &&& $\checkmark$ & 43.0 \\
&& $\checkmark$ &&& 43.2 \\
\rowcolor{gray!10}
&& $\checkmark$ && $\checkmark$ & 43.4 \\
&&& $\checkmark$ && 40.2 \\
\rowcolor{gray!10}
&&& $\checkmark$ & $\checkmark$ & 43.3 \\
\hline
\end{tabular}
\end{center}
\end{table}
\setlength{\tabcolsep}{1.4pt}

\subsection{Other feature-based distillation methods with the normalized features}
In object detection, some feature-based distillation methods \cite{zhixing2021distilling,yang2021focal} transfer the knowledge within some pre-defined imitation regions, and achieved competitive results. In contrast, we believe full map feature imitation with our PKD can already outperform them. In order to show the generality of our method, we take FitNet \cite{romero2014fitnets}, GT Mask, FRS \cite{zhixing2021distilling} and FGD \cite{yang2021focal} as baselines and build PKD on them.

\textbf{Recaps.}
Full feature map distillation with MSE loss is used in FitNet. While the GT Mask only imitates features that overlap with ground truth bounding boxes. Feature imitation with FRS uses the aggregated classification score, takes maximum operation in channel direction, of the classification head output as a weighted score mask, to guide the distillation of the FPN. And global distillation in FGD rebuilds the relation between different pixels and transfers it from teachers to students through GcBlock \cite{cao2019gcnet}.

\textbf{Results.}
For a fair comparison, all the hyper-parameters are copied from the original paper. Results in Table \ref{featurebased} show that combining PKD with these feature imitation methods can further improve their performance.

\begin{figure}[htb] 
  \begin{minipage}{0.5\textwidth} 
    \centering 
    \vspace{-3em}
    \captionof{table}{Comparisons between feature imitation with MSE and our PKD. We address the above three issues and achieve better performance.}
    \label{tab:mse}
    \resizebox{0.95\linewidth}{!}{
        \begin{tabular}{lcccc}
        \hline\noalign{\smallskip}
        Detector Pairs & Methods & $mAP$ & $AP_{S}$ & $AP_{L}$ \\ 
        \noalign{\smallskip}
        \hline
        \noalign{\smallskip}
        \multirow{2}*{\shortstack{FCOS-ResX101 \\ Retina-Res50}} & MSE & 36.3  & 20.0 & 47.1 \\
        & Ours & 41.3 & 24.2 & 55.4 \\
        \noalign{\smallskip}
        \hline
        \noalign{\smallskip}
        \multirow{2}*{\shortstack{GFL-Res101 \\ FCOS-Res50}} & MSE & 43.0 & 27.1 & 54.3 \\
        & Ours            & 43.5 & 26.5 & 55.7 \\
        \noalign{\smallskip}
        \hline
        \noalign{\smallskip}
        \multirow{2}*{\shortstack{Retina-ResX101 \\ Retina-Res50}} & MSE & 40.4 & 22.2 & 54.4 \\
        & Ours           & 40.8  & 23.0 & 54.7 \\
        \hline
        \end{tabular}
        } 
  \end{minipage}%
  \begin{minipage}{0.5\textwidth} 
    \label{fig:con}
    \centering
    \resizebox{0.98\linewidth}{!}{
        \includegraphics[width=\linewidth]{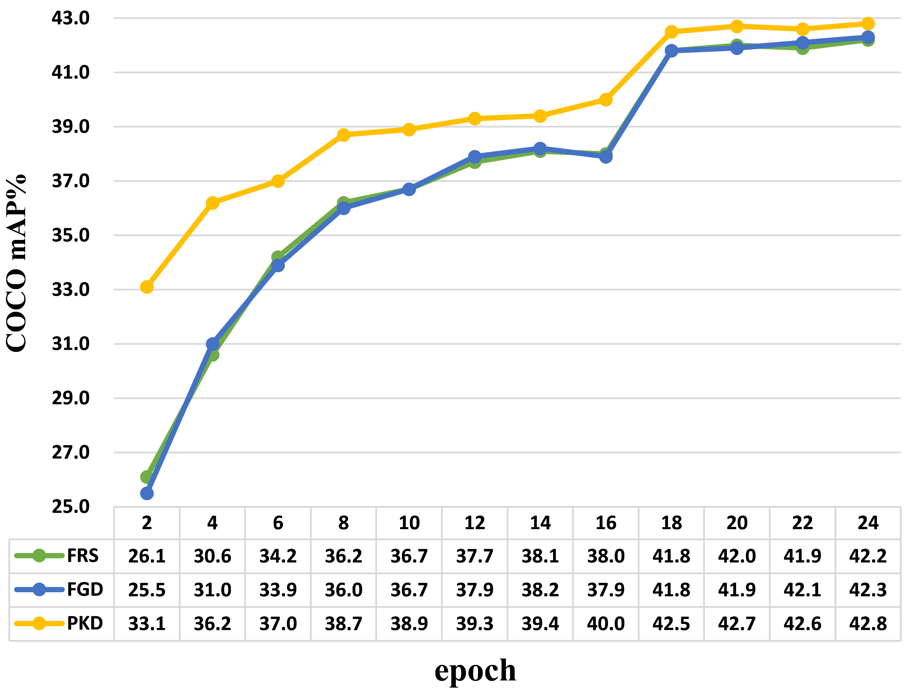}
    }
    \caption{Comparison of the convergence speed of FCOS-Res50 among other state-of-the-art distillation methods and ours.}
    \label{fig:convergence}
  \end{minipage} 
\end{figure}

\subsection{Analysis}
\subsubsection{Effectiveness of Pearson Correlation Coefficient}
Among most previous works, feature-based distillation over multi-scale features is adopted for distillation. In this study, we argue that the magnitude difference, dominant FPN stages and channels can negatively interfere with the training phase of the student and lead to sub-optimal results. To study this empirically, we conduct the following three experiments to explore the effects of MSE loss which is a widely-used loss function in distillation and suffers from the above three issues simultaneously. 
For a fair comparison, we tune the loss weight of MSE for all experiments. More details are listed in Section \ref{appendix:mse}.

For the first pair, FCOS-ResX101 is used as the teacher whose feature magnitude is significantly different from that of the student Retina-Res50 (refer to Figure \ref{fig:magnitude}). In this case, what constitutes the knowledge is better presented by relational information from the teacher's features than the absolute values. For the second pair, GFL-Res101 is the teacher and FCOS-Res50 is the student. For both teacher and student, features in FPN stages 'P5' and 'P6' are less activated than those in stages 'P3' and 'P4' (refer to Figure \ref{fig:stage}). Hence, features in stages 'P3' and 'P4' could dominate the gradient of the distillation loss, which will overwhelm the effects of other features. Since larger objects are usually assigned to higher feature levels, the student's performance for large objects is significantly lower than that of ours when MSE is used as the distillation loss. For the last pair, Retina-ResX101 is the teacher and Retina-Res50 is the student. There are always a few channels with greater values (refer to Figure \ref{fig:channel1} and Figure \ref{fig:channel2}) , so directly imitating the feature maps may introduce much noise in dominant channels.

By comparing the results in Table \ref{tab:mse}, we find that our proposed PKD addresses the above three issues and achieves better performance. Hence, an effective distillation loss function should have the ability to handle the above three problems. We hope PKD could serve as a solid baseline and help ease future research in knowledge distillation community.

\subsubsection{Convergence Speed}
In this subsection, we conduct experiments with FCOS to compare the convergence speed of our method with other state-of-the-art distillation methods on the COCO benchmark. 
As shown in Figure \ref{fig:convergence}, the training convergence can significantly speed up in the early training stage with our PKD. Meanwhile, the final performance of the student detector FCOS-Res50 is about 0.5\%-0.6\% $mAP$ higher than that of FGD and FRS.
In addition, there is no need to forward the detection head of the teacher, which reduces the training time significantly, especially for those models with cascaded heads. 

\subsubsection{Sensitivity study of loss weight $\alpha$}
\begin{wraptable}{r}{5cm}
\centering
\vspace{-1.0em}
\caption{\raggedbottom Ablation study of loss weight hyper-parameter $\alpha$ on FCOS ResX101 - RetinaNet Res50.} \label{wrap-tab:hp}
\vspace{-1.0em}
\begin{tabular}{cccccc} \\ 
\toprule  
$\alpha$ & 3 & 5 & 8 & 10 & 13 \\
\midrule
$mAP$ & 41.0 & 41.1 & 41.1 & 41.3 & 41.1 \\ 
\bottomrule
\vspace{-2.0em}
\end{tabular}
\end{wraptable}
In Eq. \ref{overall}, we use the loss weight hyper-parameter $\alpha$ to balance the detection training loss and distillation loss. Here, we conduct several experiments to investigate the influence of $\alpha$.  As shown in Table \ref{wrap-tab:hp}, the worst result is just a 0.3 $mAP$ drop compared with the best result, indicating our method is not sensitive to the only hyper-parameter $\alpha$. In addition, thanks to the normalization mechanism in PCC, a relatively uniform value of loss weight $\alpha$ can be found among different teacher-student detector pairs to keep the balance of the detection loss and the distillation loss unchanged. 

\section{Conclusion} \label{sec:conclusion}
This paper empirically finds that FPN feature imitation is applicable for heterogeneous detector pairs although their detection heads and label assignments are different. Then we propose feature imitation with Pearson Correlation Coefficient to focus on the relational information from the teacher and relax the distribution constraints of the student's feature value magnitude. Furthermore, a general KD framework is proposed, capable of distilling knowledge for both homogeneous and heterogeneous detector pairs. It converges faster and only introduces one hyper-parameter, which can easily be applied to other datasets. However, our understanding of whether our proposed PKD is capable of other tasks such as text recognition is preliminary and left as future works.

\section*{Acknowledgments}
This work is supported in part by the Shanghai Committee of Science and Technology, China (Grant No. 20DZ1100800), NSFC 62273347, the National Key Research and Development Program of China (2020AAA0103402), Jiangsu Key Research and Development Plan (No.BE2021012-2) and NSFC 61876182.

{\small
    \bibliographystyle{plain}
    \bibliography{egbib}

\begin{thebibliography}{10}

\bibitem{ahn2019variational}
Sungsoo Ahn, Shell~Xu Hu, Andreas Damianou, Neil~D Lawrence, and Zhenwen Dai.
\newblock Variational information distillation for knowledge transfer.
\newblock In {\em Proceedings of the IEEE/CVF Conference on Computer Vision and
  Pattern Recognition}, pages 9163--9171, 2019.

\bibitem{benesty2009pearson}
Jacob Benesty, Jingdong Chen, Yiteng Huang, and Israel Cohen.
\newblock Pearson correlation coefficient.
\newblock In {\em Noise reduction in speech processing}, pages 1--4. Springer,
  2009.

\bibitem{cai2018cascade}
Zhaowei Cai and Nuno Vasconcelos.
\newblock Cascade r-cnn: Delving into high quality object detection.
\newblock In {\em Proceedings of the IEEE conference on computer vision and
  pattern recognition}, pages 6154--6162, 2018.

\bibitem{cai2019cascade}
Zhaowei Cai and Nuno Vasconcelos.
\newblock Cascade r-cnn: high quality object detection and instance
  segmentation.
\newblock {\em IEEE transactions on pattern analysis and machine intelligence},
  43(5):1483--1498, 2019.

\bibitem{cao2019gcnet}
Yue Cao, Jiarui Xu, Stephen Lin, Fangyun Wei, and Han Hu.
\newblock Gcnet: Non-local networks meet squeeze-excitation networks and
  beyond.
\newblock In {\em Proceedings of the IEEE/CVF international conference on
  computer vision workshops}, pages 0--0, 2019.

\bibitem{chen2017learning}
Guobin Chen, Wongun Choi, Xiang Yu, Tony Han, and Manmohan Chandraker.
\newblock Learning efficient object detection models with knowledge
  distillation.
\newblock {\em Advances in neural information processing systems}, 30, 2017.

\bibitem{chen2019mmdetection}
Kai Chen, Jiaqi Wang, Jiangmiao Pang, Yuhang Cao, Yu~Xiong, Xiaoxiao Li,
  Shuyang Sun, Wansen Feng, Ziwei Liu, Jiarui Xu, et~al.
\newblock Mmdetection: Open mmlab detection toolbox and benchmark.
\newblock {\em arXiv preprint arXiv:1906.07155}, 2019.

\bibitem{2021mmrazor}
MMRazor Contributors.
\newblock Openmmlab model compression toolbox and benchmark.
\newblock \url{https://github.com/open-mmlab/mmrazor}, 2021.

\bibitem{dai2021general}
Xing Dai, Zeren Jiang, Zhao Wu, Yiping Bao, Zhicheng Wang, Si~Liu, and Erjin
  Zhou.
\newblock General instance distillation for object detection.
\newblock In {\em Proceedings of the IEEE/CVF Conference on Computer Vision and
  Pattern Recognition}, pages 7842--7851, 2021.

\bibitem{duan2019centernet}
Kaiwen Duan, Song Bai, Lingxi Xie, Honggang Qi, Qingming Huang, and Qi~Tian.
\newblock Centernet: Keypoint triplets for object detection.
\newblock In {\em Proceedings of the IEEE/CVF international conference on
  computer vision}, pages 6569--6578, 2019.

\bibitem{feng2021tood}
Chengjian Feng, Yujie Zhong, Yu~Gao, Matthew~R Scott, and Weilin Huang.
\newblock Tood: Task-aligned one-stage object detection.
\newblock In {\em 2021 IEEE/CVF International Conference on Computer Vision
  (ICCV)}, pages 3490--3499. IEEE Computer Society, 2021.

\bibitem{guo2021distilling}
Jianyuan Guo, Kai Han, Yunhe Wang, Han Wu, Xinghao Chen, Chunjing Xu, and Chang
  Xu.
\newblock Distilling object detectors via decoupled features.
\newblock In {\em Proceedings of the IEEE/CVF Conference on Computer Vision and
  Pattern Recognition}, pages 2154--2164, 2021.

\bibitem{he2017mask}
Kaiming He, Georgia Gkioxari, Piotr Doll{\'a}r, and Ross Girshick.
\newblock Mask r-cnn.
\newblock In {\em Proceedings of the IEEE international conference on computer
  vision}, pages 2961--2969, 2017.

\bibitem{heo2019knowledge}
Byeongho Heo, Minsik Lee, Sangdoo Yun, and Jin~Young Choi.
\newblock Knowledge transfer via distillation of activation boundaries formed
  by hidden neurons.
\newblock In {\em Proceedings of the AAAI Conference on Artificial
  Intelligence}, volume~33, pages 3779--3787, 2019.

\bibitem{hinton2015distilling}
Geoffrey Hinton, Oriol Vinyals, Jeff Dean, et~al.
\newblock Distilling the knowledge in a neural network.
\newblock {\em arXiv preprint arXiv:1503.02531}, 2(7), 2015.

\bibitem{huang2017like}
Zehao Huang and Naiyan Wang.
\newblock Like what you like: Knowledge distill via neuron selectivity
  transfer.
\newblock {\em arXiv preprint arXiv:1707.01219}, 2017.

\bibitem{ioffe2015batch}
Sergey Ioffe and Christian Szegedy.
\newblock Batch normalization: Accelerating deep network training by reducing
  internal covariate shift.
\newblock In {\em International conference on machine learning}, pages
  448--456. PMLR, 2015.

\bibitem{kang2021instance}
Zijian Kang, Peizhen Zhang, Xiangyu Zhang, Jian Sun, and Nanning Zheng.
\newblock Instance-conditional knowledge distillation for object detection.
\newblock {\em Advances in Neural Information Processing Systems}, 34, 2021.

\bibitem{li2021knowledge}
Gang Li, Xiang Li, Yujie Wang, Shanshan Zhang, Yichao Wu, and Ding Liang.
\newblock Knowledge distillation for object detection via rank mimicking and
  prediction-guided feature imitation.
\newblock {\em arXiv preprint arXiv:2112.04840}, 2021.

\bibitem{li2017mimicking}
Quanquan Li, Shengying Jin, and Junjie Yan.
\newblock Mimicking very efficient network for object detection.
\newblock In {\em Proceedings of the ieee conference on computer vision and
  pattern recognition}, pages 6356--6364, 2017.

\bibitem{li2020generalized}
Xiang Li, Wenhai Wang, Lijun Wu, Shuo Chen, Xiaolin Hu, Jun Li, Jinhui Tang,
  and Jian Yang.
\newblock Generalized focal loss: Learning qualified and distributed bounding
  boxes for dense object detection.
\newblock {\em Advances in Neural Information Processing Systems},
  33:21002--21012, 2020.

\bibitem{lin2017feature}
Tsung-Yi Lin, Piotr Doll{\'a}r, Ross Girshick, Kaiming He, Bharath Hariharan,
  and Serge Belongie.
\newblock Feature pyramid networks for object detection.
\newblock In {\em Proceedings of the IEEE conference on computer vision and
  pattern recognition}, pages 2117--2125, 2017.

\bibitem{lin2017focal}
Tsung-Yi Lin, Priya Goyal, Ross Girshick, Kaiming He, and Piotr Doll{\'a}r.
\newblock Focal loss for dense object detection.
\newblock In {\em Proceedings of the IEEE international conference on computer
  vision}, pages 2980--2988, 2017.

\bibitem{lin2014microsoft}
Tsung-Yi Lin, Michael Maire, Serge Belongie, James Hays, Pietro Perona, Deva
  Ramanan, Piotr Doll{\'a}r, and C~Lawrence Zitnick.
\newblock Microsoft coco: Common objects in context.
\newblock In {\em European conference on computer vision}, pages 740--755.
  Springer, 2014.

\bibitem{liu2016ssd}
Wei Liu, Dragomir Anguelov, Dumitru Erhan, Christian Szegedy, Scott Reed,
  Cheng-Yang Fu, and Alexander~C Berg.
\newblock Ssd: Single shot multibox detector.
\newblock In {\em European conference on computer vision}, pages 21--37.
  Springer, 2016.

\bibitem{liu2019knowledge}
Yufan Liu, Jiajiong Cao, Bing Li, Chunfeng Yuan, Weiming Hu, Yangxi Li, and
  Yunqiang Duan.
\newblock Knowledge distillation via instance relationship graph.
\newblock In {\em Proceedings of the IEEE/CVF Conference on Computer Vision and
  Pattern Recognition}, pages 7096--7104, 2019.

\bibitem{liu2021Swin}
Ze~Liu, Yutong Lin, Yue Cao, Han Hu, Yixuan Wei, Zheng Zhang, Stephen Lin, and
  Baining Guo.
\newblock Swin transformer: Hierarchical vision transformer using shifted
  windows.
\newblock {\em arXiv preprint arXiv:2103.14030}, 2021.

\bibitem{park2019relational}
Wonpyo Park, Dongju Kim, Yan Lu, and Minsu Cho.
\newblock Relational knowledge distillation.
\newblock In {\em Proceedings of the IEEE/CVF Conference on Computer Vision and
  Pattern Recognition}, pages 3967--3976, 2019.

\bibitem{paszke2017automatic}
Adam Paszke, Sam Gross, Soumith Chintala, Gregory Chanan, Edward Yang, Zachary
  DeVito, Zeming Lin, Alban Desmaison, Luca Antiga, and Adam Lerer.
\newblock Automatic differentiation in pytorch.
\newblock 2017.

\bibitem{redmon2018yolov3}
Joseph Redmon and Ali Farhadi.
\newblock Yolov3: An incremental improvement.
\newblock {\em arXiv preprint arXiv:1804.02767}, 2018.

\bibitem{ren2015faster}
Shaoqing Ren, Kaiming He, Ross Girshick, and Jian Sun.
\newblock Faster r-cnn: Towards real-time object detection with region proposal
  networks.
\newblock {\em Advances in neural information processing systems}, 28, 2015.

\bibitem{romero2014fitnets}
Adriana Romero, Nicolas Ballas, Samira~Ebrahimi Kahou, Antoine Chassang, Carlo
  Gatta, and Yoshua Bengio.
\newblock Fitnets: Hints for thin deep nets.
\newblock {\em arXiv preprint arXiv:1412.6550}, 2014.

\bibitem{shu2021channel}
Changyong Shu, Yifan Liu, Jianfei Gao, Zheng Yan, and Chunhua Shen.
\newblock Channel-wise knowledge distillation for dense prediction.
\newblock In {\em Proceedings of the IEEE/CVF International Conference on
  Computer Vision}, pages 5311--5320, 2021.

\bibitem{sun2020distilling}
Ruoyu Sun, Fuhui Tang, Xiaopeng Zhang, Hongkai Xiong, and Qi~Tian.
\newblock Distilling object detectors with task adaptive regularization.
\newblock {\em arXiv preprint arXiv:2006.13108}, 2020.

\bibitem{tian2019contrastive}
Yonglong Tian, Dilip Krishnan, and Phillip Isola.
\newblock Contrastive representation distillation.
\newblock {\em arXiv preprint arXiv:1910.10699}, 2019.

\bibitem{tian2019fcos}
Zhi Tian, Chunhua Shen, Hao Chen, and Tong He.
\newblock Fcos: Fully convolutional one-stage object detection.
\newblock In {\em Proceedings of the IEEE/CVF international conference on
  computer vision}, pages 9627--9636, 2019.

\bibitem{tung2019similarity}
Frederick Tung and Greg Mori.
\newblock Similarity-preserving knowledge distillation.
\newblock In {\em Proceedings of the IEEE/CVF International Conference on
  Computer Vision}, pages 1365--1374, 2019.

\bibitem{ulyanov2016instance}
Dmitry Ulyanov, Andrea Vedaldi, and Victor Lempitsky.
\newblock Instance normalization: The missing ingredient for fast stylization.
\newblock {\em arXiv preprint arXiv:1607.08022}, 2016.

\bibitem{wang2019distilling}
Tao Wang, Li~Yuan, Xiaopeng Zhang, and Jiashi Feng.
\newblock Distilling object detectors with fine-grained feature imitation.
\newblock In {\em Proceedings of the IEEE/CVF Conference on Computer Vision and
  Pattern Recognition}, pages 4933--4942, 2019.

\bibitem{wang2020intra}
Yukang Wang, Wei Zhou, Tao Jiang, Xiang Bai, and Yongchao Xu.
\newblock Intra-class feature variation distillation for semantic segmentation.
\newblock In {\em European Conference on Computer Vision}, pages 346--362.
  Springer, 2020.

\bibitem{wu2018group}
Yuxin Wu and Kaiming He.
\newblock Group normalization.
\newblock In {\em Proceedings of the European conference on computer vision
  (ECCV)}, pages 3--19, 2018.

\bibitem{yang2019reppoints}
Ze~Yang, Shaohui Liu, Han Hu, Liwei Wang, and Stephen Lin.
\newblock Reppoints: Point set representation for object detection.
\newblock In {\em Proceedings of the IEEE/CVF International Conference on
  Computer Vision}, pages 9657--9666, 2019.

\bibitem{yang2021focal}
Zhendong Yang, Zhe Li, Xiaohu Jiang, Yuan Gong, Zehuan Yuan, Danpei Zhao, and
  Chun Yuan.
\newblock Focal and global knowledge distillation for detectors.
\newblock {\em arXiv preprint arXiv:2111.11837}, 2021.

\bibitem{yao2021g}
Lewei Yao, Renjie Pi, Hang Xu, Wei Zhang, Zhenguo Li, and Tong Zhang.
\newblock G-detkd: Towards general distillation framework for object detectors
  via contrastive and semantic-guided feature imitation.
\newblock In {\em Proceedings of the IEEE/CVF International Conference on
  Computer Vision}, pages 3591--3600, 2021.

\bibitem{yim2017gift}
Junho Yim, Donggyu Joo, Jihoon Bae, and Junmo Kim.
\newblock A gift from knowledge distillation: Fast optimization, network
  minimization and transfer learning.
\newblock In {\em Proceedings of the IEEE Conference on Computer Vision and
  Pattern Recognition}, pages 4133--4141, 2017.

\bibitem{zhang2020improve}
Linfeng Zhang and Kaisheng Ma.
\newblock Improve object detection with feature-based knowledge distillation:
  Towards accurate and efficient detectors.
\newblock In {\em International Conference on Learning Representations}, 2020.

\bibitem{zhao2022decoupled}
Borui Zhao, Quan Cui, Renjie Song, Yiyu Qiu, and Jiajun Liang.
\newblock Decoupled knowledge distillation.
\newblock {\em arXiv preprint arXiv:2203.08679}, 2022.

\bibitem{zhixing2021distilling}
Du~Zhixing, Rui Zhang, Ming Chang, Shaoli Liu, Tianshi Chen, Yunji Chen, et~al.
\newblock Distilling object detectors with feature richness.
\newblock {\em Advances in Neural Information Processing Systems}, 34, 2021.

\end{thebibliography}
}


\newpage
\appendix

\section{Appendix}

\subsection{Motivation to build a general KD framework}
Modern detectors are roughly divided into two-stage detectors \cite{ren2015faster,he2017mask,cai2018cascade} and dense prediction detectors ({\it e.g.}, anchor-based one-stage detectors \cite{lin2017focal,liu2016ssd,redmon2018yolov3} and anchor-free one-stage detectors \cite{duan2019centernet,tian2019fcos,yang2019reppoints}).
Each family has its own advantages and weakness. In particular, two-stage detectors usually have higher performance, while being slower in inference speed and harder to be deployed due to the region proposal network (RPN) and RCNN head. On the other hand, dense prediction detectors are faster than two-stage detectors while being less accurate. 
In practice, it is a natural idea to use two-stage detectors as teachers to enhance dense prediction detectors.

Moreover, knowledge distillation between heterogeneous dense prediction detector pairs is also promising. In some scenarios, only the detectors with a specific architecture can be deployed due to hardware limitations. For example, compared with Batch Normalization \cite{ioffe2015batch} and Instance Normalization \cite{ulyanov2016instance}, Group Normalization \cite{wu2018group} is hard to deploy. However, the most powerful teachers may belong to different categories.
Furthermore, object detection is developing rapidly and algorithms with better performance are proposed continuously. Nevertheless, it is not easy to change detectors frequently in terms of stability and hardware runtime limitations in practical applications. So it is helpful if knowledge distillation can be conducted between the latest high-capacity detectors and the widely-used compact detectors. 

Hence, we are motivated to design a general distillation method capable of distilling knowledge between both homogeneous and heterogeneous detector pairs.

\subsection{Details of Training Recipe} \label{appendix:details}
We conduct experiments on different detection frameworks, including two-stage models, anchor-based one-stage models and anchor-free one-stage models. \cite{kang2021instance} proposes inheriting strategy which initializes the student with the teacher’s neck and head parameters and gets better results. Here we use this strategy to initialize the student which has the same head structure as the teacher and find that it helps students converge faster. 

All experiments are performed on 8 Tesla A100 GPUs with 2 images in each. Our implementation is based on mmdetection \cite{chen2019mmdetection} and mmrazor \cite{2021mmrazor} with Pytorch framework \cite{paszke2017automatic}. '1x' (namely 12 epochs), '2x' (namely 24 epochs) and '2x+ms' (namely 24 epochs with multi-scale training) training schedules with SGD optimizer are used. Momentum and weight decay are set to 0.9 and 1e-4. The initial learning rate is set to 0.02 for Faster RCNN and 0.01 for others. We train FCOS \cite{tian2019fcos} with tricks including GIoULoss, norm-on-bbox and center-sampling which is the same as FGD \cite{yang2021focal} and GID \cite{dai2021general}.
For distillation, only one hyper-parameter $\alpha$ is introduced to balance the supervised learning loss and distillation loss, and it is set to 6 when using a two-stage detector as the teacher and 10 when using a one-stage one as the teacher. The teacher network is well-trained previously and fixed during training.

\subsection{Connection of PCC and KL divergence} \label{append:connection}
Section \ref{sec:kl} in the main text shows the connection between PCC and KL divergence. We conduct two experiments on RetinaNet and GFL to verify this empirically. For both experiments, we set loss weight $\alpha = 10$ and temperature $T = 50$. As shown in Table \ref{table:normkl}, minimizing KL divergence between post-normalized features in the high-temperature limit can also achieve similar results.

\subsection{Details of feature imitation with MSE} \label{appendix:mse}
As shown in Table \ref{tab:mse} in the main text, we compare the results of MSE with our proposed PKD. As the gradient of MSE loss depends on feature value magnitude and it is usually different among different detectors, we have to tune the loss weight carefully to achieve relatively good results, as shown in Table \ref{table:appendix_mse}. And we put the best results down in Table \ref{tab:mse}.

\setlength{\tabcolsep}{4pt}
\begin{table}[htpb]
\begin{center}
\caption{Results of the KL divergence with normalization mechanism.}
\label{table:normkl}
\begin{tabular}{lclccccc}
\hline\noalign{\smallskip}
Method & schedule & $mAP$ & $AP_{50}$ & $AP_{75}$ & $AP_{S}$ & $AP_{M}$ & $AP_{L}$  \\
\noalign{\smallskip}
\hline
\noalign{\smallskip}
Retina-ResX101 (T) & 2x & 40.8 & 60.5 & 43.7 & 22.9 & 44.5 & 54.6 \\
Retina-Res50   (S) & 2x & 37.4 & 56.7 & 39.6 & 20.0 & 40.7 & 49.7 \\
Norm+KL            & 2x & 40.9 & 60.3 & 43.6 & 22.9 & 45.2 & 55.1 \\
\noalign{\smallskip}
\hline
\noalign{\smallskip}
GFL-Res101 (T)    & 2x+ms & 44.9 & 63.1 & 49.0 & 28.0 & 49.1 & 57.2 \\
GFL-Res50 (S)     & 1x    & 40.2 & 58.4 & 43.3 & 23.3 & 44.0 & 52.2 \\
Norm+KL           & 1x    & 43.1 & 60.9 & 46.7 & 25.1 & 47.8 & 55.9 \\
\hline
\end{tabular}
\end{center}
\end{table}
\setlength{\tabcolsep}{1.4pt}

\setlength{\tabcolsep}{4pt}
\begin{table}[htpb]
\begin{center}
\caption{Results of MSE on various detector pairs.}
\label{table:appendix_mse}
\begin{tabular}{cccccc}
\hline\noalign{\smallskip}
Teacher & Student & Schedule & Baseline & Loss Weight & $mAP$  \\
\noalign{\smallskip}
\hline
\noalign{\smallskip}
\multirow{3}*{FCOS-ResX101} & \multirow{3}*{Retina-Res50} & \multirow{3}*{1x} & \multirow{3}*{36.5} & 5 & 33.9 \\
& & & & 10 & 31.4 \\
& & & & 20 & 29.7 \\
\noalign{\smallskip}
\hline
\noalign{\smallskip}
\multirow{3}*{FCOS-ResX101} & \multirow{3}*{Retina-Res50} & \multirow{3}*{2x} & \multirow{3}*{37.4} & 5 & 36.3 \\
& & & & 10 & 35.6 \\
& & & & 20 & 34.9 \\
\noalign{\smallskip}
\hline
\noalign{\smallskip}
\multirow{2}*{GFL-Res101} & \multirow{2}*{FCOS-Res50} & \multirow{2}*{1x} & \multirow{2}*{36.6} & 50 & 38.7 \\
& & & & 70 & 39.2 \\
\noalign{\smallskip}
\hline
\noalign{\smallskip}
\multirow{5}*{GFL-Res101} & \multirow{5}*{FCOS-Res50} & \multirow{5}*{2x} & \multirow{5}*{39.1} & 10 & 41.5 \\
& & & & 30 & 42.5 \\
& & & & 50 & 42.7 \\
& & & & 70 & 43.0 \\
& & & & 80 & 42.9 \\
\noalign{\smallskip}
\hline
\noalign{\smallskip}
\multirow{3}*{Retina-ResX101} & \multirow{3}*{Retina-Res50} & \multirow{3}*{2x} & \multirow{3}*{37.4} & 5 & 40.0 \\
& & & & 10 & 40.4 \\
& & & & 15 & 40.3 \\
\noalign{\smallskip}
\hline
\noalign{\smallskip}
\multirow{4}*{MaskRCNN-Swin} & \multirow{4}*{FasterRCNN-Res50} & \multirow{4}*{2x} & \multirow{4}*{38.4} & 3 & 41.7 \\
& & & & 5 & 41.7 \\
& & & & 6 & 41.6 \\
& & & & 10 & 41.7 \\
\hline
\end{tabular}
\end{center}
\end{table}
\setlength{\tabcolsep}{1.4pt}

\subsection{Effectiveness of Pearson Correlation Coefficient}
\begin{figure}[htpb]
\centering
\includegraphics[width=\linewidth]{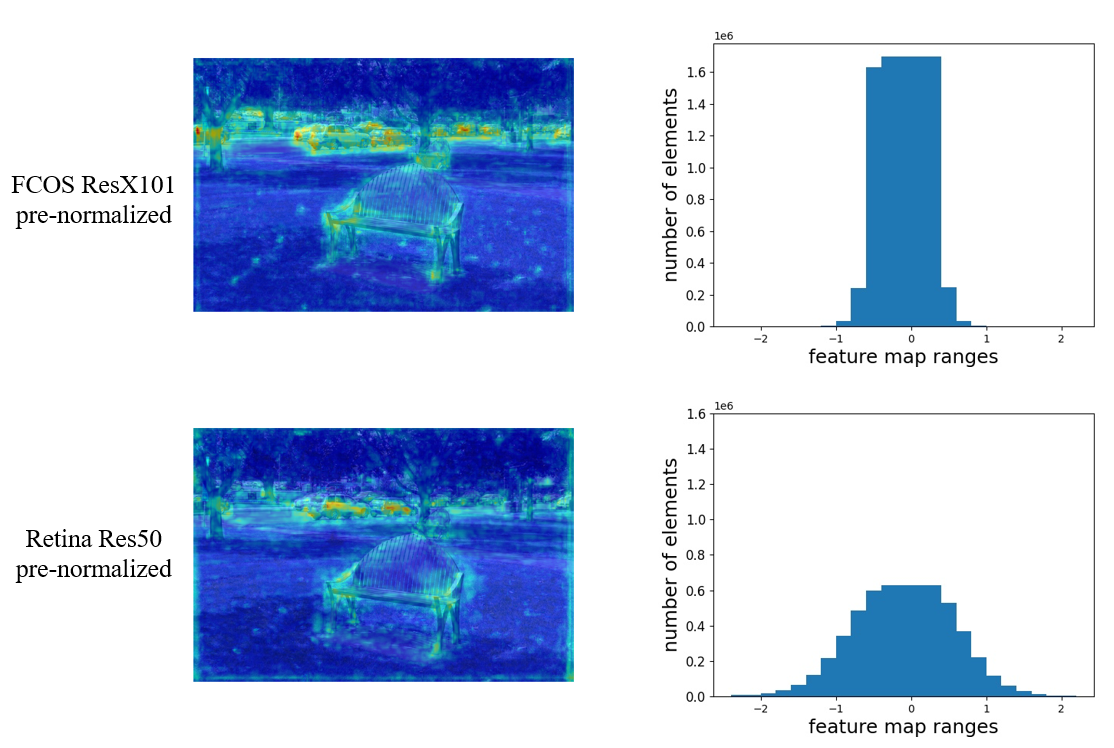}
\caption{Visualization of the activation patterns and activation distribution of FPN stage 'P3'.}
\label{fig:magnitude}
\end{figure}

\begin{figure}
\centering
\includegraphics[width=\linewidth]{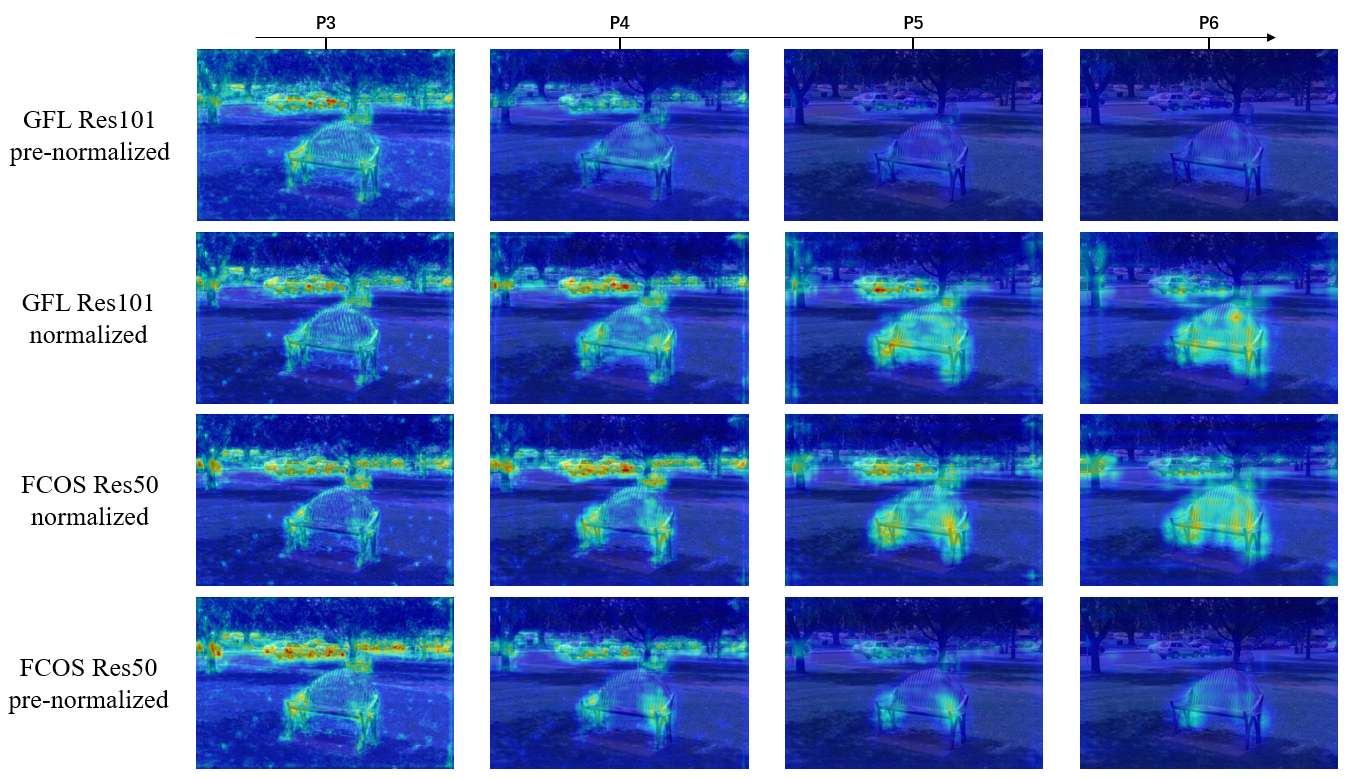}
\caption{Visualization of dominant FPN stages. From left to right, they correspond to the activation patterns in FPN stage 'P3' to 'P6'. The leftmost corresponds to the lowest stage of FPN, and the rightmost corresponds to the highest stage of FPN.}
\label{fig:stage}
\end{figure}

\begin{figure}
\centering
\includegraphics[width=\linewidth]{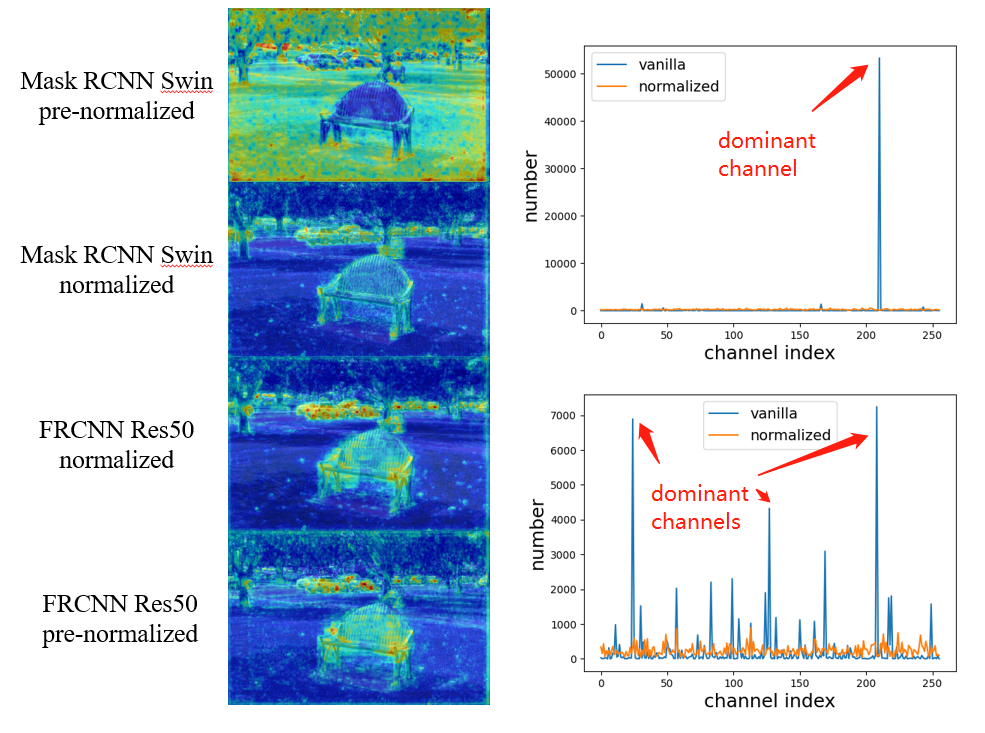}
\caption{Visualization of dominant channels in MaskRCNN-Swin and FasterRCNN-Res50. Left: Visualization of the activation patterns of FPN stage 'P3'. Right: Dominant channels in pre-normalized FPN stage 'P3'. Let $\bm{s_{l,u,v}} \in \mathbb{R}^{C}$ denote the feature vector located in pixel (u, v) from $l$-th FPN stage and omit $l$ for clarity. Then $number_i = \sum_{u, v} \mathbbm{1} [\arg \max_{c} s_{u,v}^{(c)} = i]$ where $i$ is the channel index. We define channels with a larger $number$ as dominant channels.}
\label{fig:channel1}
\end{figure}

\begin{figure}
\centering
\includegraphics[width=\linewidth]{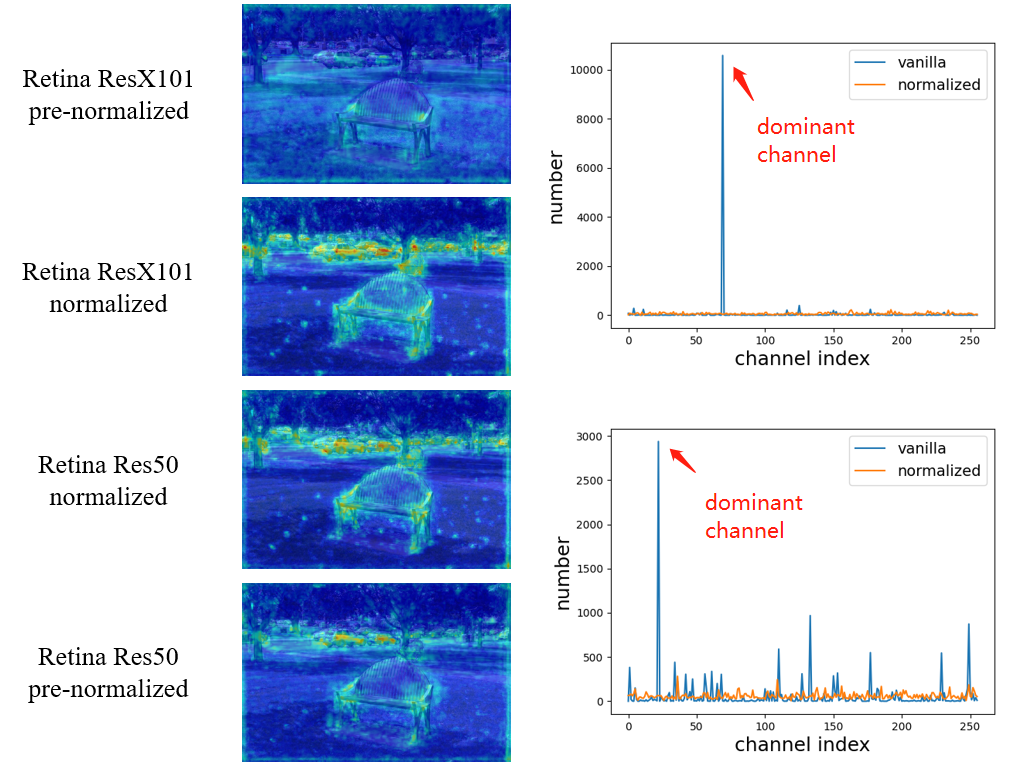}
\caption{Visualization of dominant channels in Retina-ResX101 and Retina-Res50.}
\label{fig:channel2}
\end{figure}

In this paper, we argue that the magnitude difference, dominant FPN stages and channels can negatively interfere with the training phase of the student. To clearly show that these three issues do exist, we elaborately visualize the FPN feature responses of the teacher and student detectors before distillation, as shown in Figure \ref{fig:magnitude}, Figure \ref{fig:stage}, Figure \ref{fig:channel1} and Figure \ref{fig:channel2}. We follow the visualization method in Section \ref{sec:intro}.

Through these comparisons, we obtain the following three observations:

(1) The feature value magnitude of different detectors could be significantly different, especially for heterogeneous detectors, as shown in Figure \ref{fig:magnitude}. Directly aligning the feature maps between the teacher and the student may enforce overly strict constraints and do harm to the student (see Table \ref{tab:mse}  top in the main text). 

(2) As shown in Figure \ref{fig:stage}, compared with features in FPN stage 'P3' and 'P4', features in stage 'P5' and 'P6' are less activated in both GFL-Res101 and FCOS-Res50. And it is a common case among different detector pairs. FPN stages with larger values could dominate the gradient of the distillation loss, which will overwhelm the effects of other features in KD.

(3) As shown in Figure \ref{fig:channel1} (right), the feature magnitude of the $210$-th channel of MaskRCNN-Swin FPN stage 'P6' is significantly larger than others. Similar phenomena also exist in other detectors such as RetinaNet (see Figure \ref{fig:channel2}). The small gradients produced by those less activated channels can be drowned in the large gradients produced by dominant ones, thus limiting further refinement. Furthermore, there is much noise in the object-irrelevant area, as depicted in Figure \ref{fig:channel1} (left). Directly imitating the feature maps may introduce much noise. 

Comparing Table \ref{table:appendix_mse} and Table \ref{tab:mse} in the main text, we find that our proposed PKD addresses the above three issues and achieves better performance. Hence, an effective distillation loss function should have the ability to handle the above three problems. We hope PKD could serve as a solid baseline and help ease future research in knowledge distillation community.

\begin{figure}
\centering
\includegraphics[width=\linewidth]{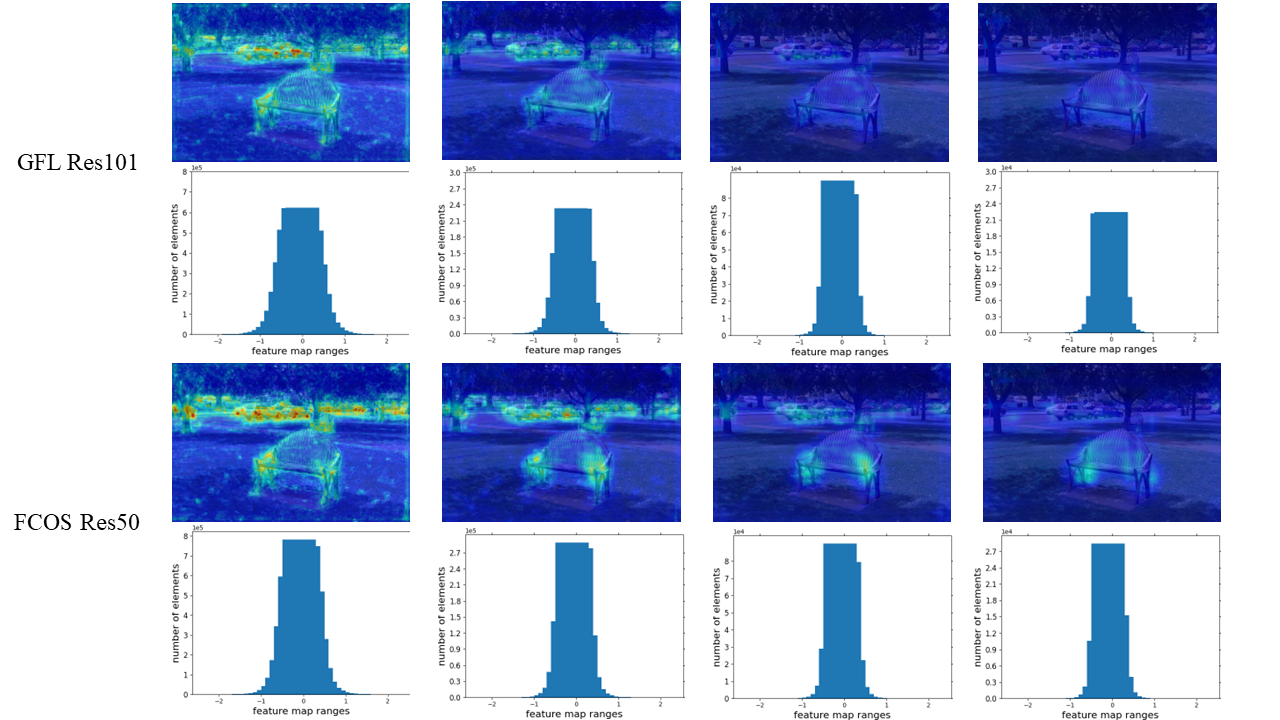}
\caption{Visualization of the activation patterns and activation distribution of GFL-Res101 and FCOS-Res50. From left to right, they correspond to the activation patterns and activation distribution in FPN stage 'P3' to 'P6'. The leftmost corresponds to the lowest stage of FPN, and the rightmost corresponds to the highest stage of FPN. The feature value magnitude of GFL and FCOS is similar.}
\end{figure}

\end{document}